\documentclass{article}

\usepackage{microtype}
\usepackage{graphicx}
\usepackage{subfigure}
\usepackage{booktabs} % for professional tables
\usepackage{array}
\usepackage{bbm}

\usepackage{hyperref}

\usepackage[accepted]{icml2024}

\usepackage{amsmath}
\usepackage{amssymb}
\usepackage{mathtools}
\usepackage{amsthm}
\usepackage{subcaption}
\usepackage{subfigure}
\usepackage[capitalize,noabbrev]{cleveref}

\usepackage[textsize=tiny]{todonotes}

\begin{document}

\twocolumn[
\icmltitle{Constructing Domain-Specific Evaluation Sets for LLM-as-a-judge}

% It is OKAY to include author information, even for blind
% submissions: the style file will automatically remove it for you
% unless you've provided the [accepted] option to the icml2024
% package.

% List of affiliations: The first argument should be a (short)
% identifier you will use later to specify author affiliations
% Academic affiliations should list Department, University, City, Region, Country
% Industry affiliations should list Company, City, Region, Country

% You can specify symbols, otherwise they are numbered in order.
% Ideally, you should not use this facility. Affiliations will be numbered
% in order of appearance and this is the preferred way.
\icmlsetsymbol{equal}{*}

\begin{icmlauthorlist}
\icmlauthor{Ravi Raju}{comp}
\icmlauthor{Swayambhoo Jain}{comp}
\icmlauthor{Bo Li}{comp}
\icmlauthor{Jonathan Li}{comp}
\icmlauthor{Urmish Thakker}{comp}
%\icmlauthor{}{sch}
%\icmlauthor{}{sch}
%\icmlauthor{}{sch}
\end{icmlauthorlist}

\icmlaffiliation{comp}{SambaNova Systems}

\icmlcorrespondingauthor{Ravi Raju}{ravi.raju@sambanovasystems.com}

\icmlkeywords{Machine Learning, Benchmarking LLMs}

\vskip 0.3in
]

% this must go after the closing bracket ] following \twocolumn[ ...

% This command actually creates the footnote in the first column
% listing the affiliations and the copyright notice.
% The command takes one argument, which is text to display at the start of the footnote.
% The \icmlEqualContribution command is standard text for equal contribution.
% Remove it (just {}) if you do not need this facility.

%\printAffiliationsAndNotice{}  % leave blank if no need to mention equal contribution
\printAffiliationsAndNotice{} % otherwise use the standard text.

\begin{abstract}
Large Language Models (LLMs) have revolutionized the landscape of machine learning, yet current benchmarks often fall short in capturing the diverse behavior of these models in real-world applications. A benchmark's usefulness is determined by its ability to clearly differentiate between models of varying capabilities (separability) and closely align with human preferences. Existing frameworks like Alpaca-Eval 2.0 LC \cite{dubois2024lengthcontrolledalpacaevalsimpleway} and Arena-Hard v0.1 \cite{li2024crowdsourced} are limited by their focus on general-purpose queries and lack of diversity across domains such as law, medicine, and multilingual contexts. In this paper, we address these limitations by introducing a novel data pipeline that curates diverse, domain-specific evaluation sets tailored for LLM-as-a-Judge frameworks. Our approach leverages a combination of manual curation, semi-supervised learning to generate clusters, and stratified sampling to ensure balanced representation across a wide range of domains and languages. The resulting evaluation set, which includes 1573 samples across 14 categories, demonstrates high separability (84\%) across ten top-ranked models, and agreement (84\%) with Chatbot Arena and (0.915) Spearman correlation. The agreement values are 9\% better than Arena Hard and 20\% better than AlpacaEval 2.0 LC, while the Spearman coefficient is 0.7 more than the next best benchmark, showcasing a significant improvement in the usefulness of the benchmark. We further provide an open-source evaluation tool that enables fine-grained analysis of model performance across user-defined categories, offering valuable insights for practitioners. This work contributes to the ongoing effort to enhance the transparency, diversity, and effectiveness of LLM evaluation methodologies.
\end{abstract}
\begin{figure}[t]
    \centering    \includegraphics[width=1.0\linewidth]{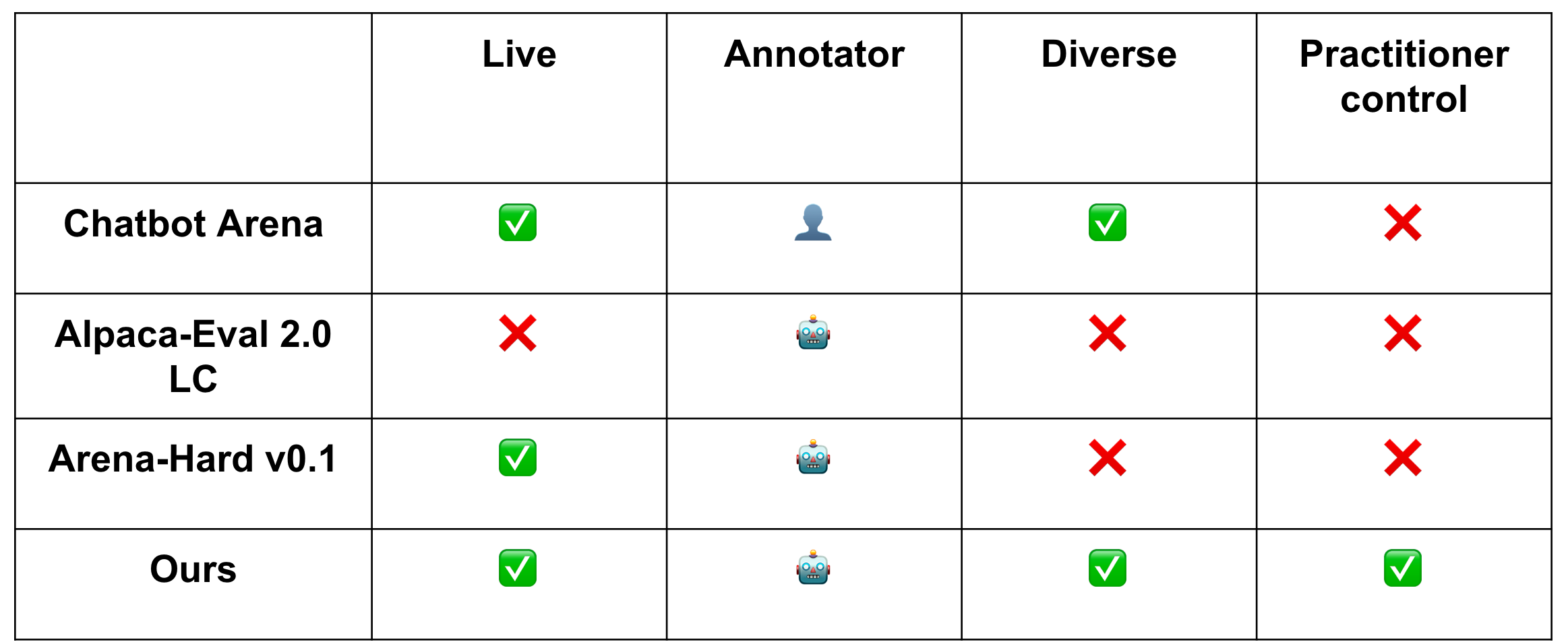}
    \caption{Compared to other benchmark frameworks our approach introduces a data pipeline that curates unlabeled data into categories that contain domains/capabilities that the practitioner cares about. It has the capability to be refreshed with new data and is diverse compared to alternatives.}
    \label{fig:motivation}
\end{figure}

\section{Introduction}
Large Language Models (LLMs) have dramatically changed the landscape of machine learning research and have been incorporated in products for the past few years. Along with their rise, a multitude of benchmarks and frameworks \cite{liang2023holisticevaluationlanguagemodels} have been proposed to assess the capabilities of LLMs which include knowledge tasks such as MMLU \cite{hendrycks2021measuringmassivemultitasklanguage}, reasoning tasks like GSM8k \cite{cobbe2021trainingverifierssolvemath} and more standard NLP tasks \cite{zellers2019hellaswagmachinereallyfinish, narayan2018dontdetailsjustsummary}. However, these benchmarks fail to capture the behavior that a user experiences in a chat/generative applications. Typically, human evaluations are seen as a gold standard to determining which LLM responses are preferable over others in a chat setting but is time-consuming and expensive to conduct \cite{chiang2024chatbotarenaopenplatform}.

To address this shortcoming, Zheng \emph{et al}. introduced the concept of LLM as a judge as an automatic evaluator alternative, which uses another LLM  the judging of model completions to another LLM such as GPT-4 or GPT-4o \cite{zheng2023judgingllmasajudgemtbenchchatbot, openai2024gpt4technicalreport}. Alpaca-Eval is another benchmark designed under the paradigm of LLM as an evaluator where a target LLM's completions are compared against a reference LLM's output (the default being GPT-4 Turbo) and assigned a winrate against the reference \cite{alpaca_eval}. It has seen widespread adoption since it is cheap, fast, and mitigates length bias \cite{chiang2024chatbotarenaopenplatform}. Similarly, Arena-Hard v0.1 is recent benchmark which focuses on distilling the Hard category of Chatbot Arena into a smaller evaluation set \cite{li2024crowdsourced}. They use a topic clustering pipeline to cluster prompts with OpenAI's embedding model (text-embedding-3-small) \cite{OpenAI2024NewEmbeddingModels} and score each cluster based on difficulty, creativity, and reasoning ability  with GPT-3.5 Turbo. They also introduce a notion of separability (how well can a benchmark differentiate between models) and agreement with human preferences (i.e. ChatBot Arena) as a measure of benchmark quality.

\begin{figure}
    \centering
    \includegraphics[width=1.0\linewidth]{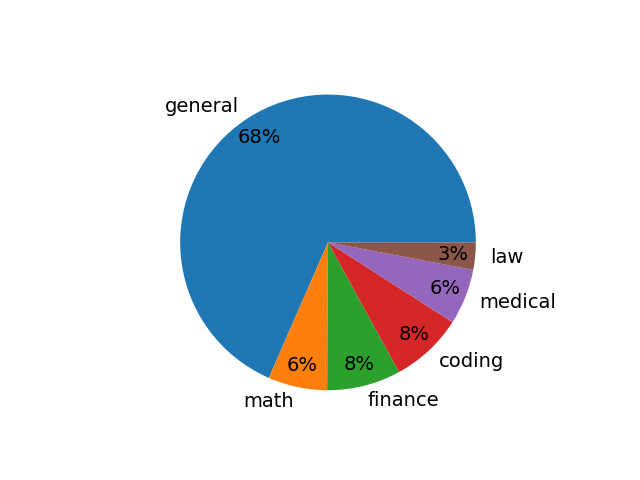}
    \caption{Alpaca-Eval category breakdown}
    \label{fig:alpaca_eval_pie_chart}
\end{figure}

Unfortunately, there are still some limitations with the current open-source LLM-as-a-judge framework. Alpaca-Eval 2.0 LC is dominated by general chat queries/instructions and has few prompts in domains such as coding, medical, finance, law and mathematics as shown in Figure \ref{fig:alpaca_eval_pie_chart}. Arena-Hard v0.1 addresses some of these deficiencies by upweighting coding and mathematics prompts and restricting the general chat queries to 30\% if the evaluation set. However, both evaluation sets are strictly in English therefore not accessing the model's multilingual capability and have a smaller number of prompts in more niche categories like law and medicine. As models are acquiring more capabilities across various data types such as charts/tables, domains and languages, it becomes crucial to determine how to evaluate each model's ability in a scalable manner. 

\begin{figure}
    \centering
    \includegraphics[width=1.0\linewidth]{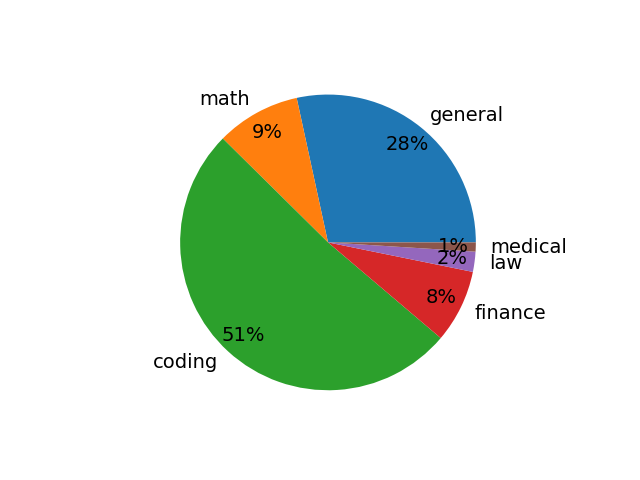}
    \caption{Arena-Hard v0.1 category breakdown}
    \label{fig:arena-hard_v0.1_pie_chart}
\end{figure}

In this paper, we attempt to address challenges from Alpaca-Eval 2.0 LC and Arena-Hard v0.1 by introducing more diversity across domain knowledge and languages. To accomplish this, we introduce a simple data pipeline methodology to create a new evaluation set designed for these specific contexts. First, we source prompts from various open source datasets (shown in Table \ref{tab:Dataset Sources}) to ensure our evaluation set has high data diversity. For the next step, we generate embeddings from a subsample of each of these datasets using an embedding model. To label the corresponding embeddings, we manually curate a seed set of prompts and label them to human-defined specific categories, generate those embeddings and train a k-NN classifier which we can use to classify the unlabeled data that we sampled. In order to make sure that no cluster/category dominates, we employ stratified sampling to ensure balanced representation across all domains and languages in the evaluation set. We further refine the quality of the prompts by manual curation and  ensure that each category has a sufficient number of prompts to mitigate the inherent variability in LLM-as-a-judge and ultimately end up with 1573 samples in the evaluation set. 

\begin{figure}
    \centering
    \includegraphics[width=1.0\linewidth]{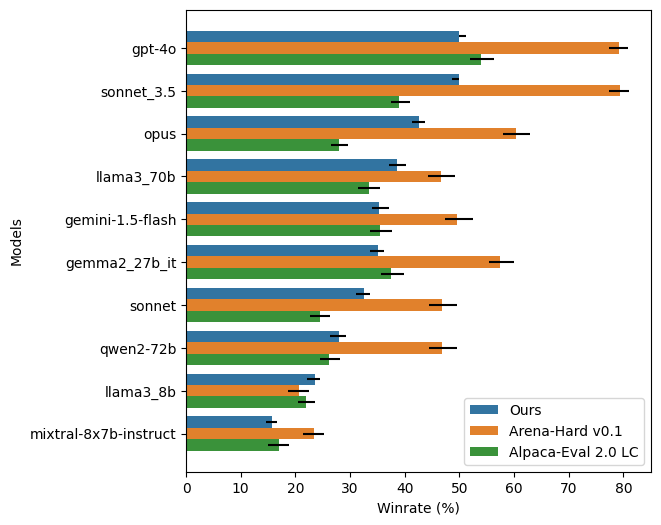}
    \caption{Visual comparison between our method, Arena-Hard v0.1, and Alpaca-Eval 2.0 LC on 10 models on separability of winrates. Our method has fewer overlaps of confidence intervals than the other baselines.}
    \label{fig:separability_barplot}
\end{figure}

There are several advantages to our approach as shown in Figure \ref{fig:motivation}. Similar to Arena-Hard v0.1, our approach is robust to contamination as we can periodically run our data pipeline on the same data to get new samples or potentially even a new data mixture. As mentioned earlier, our methodology allows introduction of new datasets which enables diversity rather than offered by Arena-Hard v0.1 and Alpaca-Eval. In addition, our evaluation set more closely mirrors Chatbot Arena rankings; Figure \ref{fig:separability_barplot} shows a visual comparison of model rankings. In particular, our evaluation set places Gemini-1.5-Flash \cite{DeepMind2024GeminiFlash} over Gemma2 27B Instruct \cite{gemma_2024} which aligns with ChatBot Arena rankings whereas the others rank Gemma2 27B over Gemini-1.5-Flash. Moreover, since we use open source models for the entire pipeline, practitioners can mold the pipeline and generate evaluation sets to test domains and capabilities they care about.

After we have obtained the evaluation set, we execute the same procedure as LLM-as-a-Judge by generating the outputs completions from GPT-4o and using them as reference to construct a leaderboard from ten various open and closed-sourced models. With this labeling approach, we are able to breakdown the composition of prompts into various categories and report category win rates. We release an evaluation tool which displays the category winrate for all models on the leaderboard and an explorer which displays both the target model as well as the reference model's completions for a prompt and the reasoning given by the LLM judge. This analysis tool allows users to obtain fine-grained insights on where different models succeed and fail for their particular use-case.

Our main contributions can be summarized below:
\begin{itemize}
  \item We introduce a new methodology that enables creation of a benchmark that tests for diverse skill sets of models. We open-source our evaluation infrastructure so practitioners can view how different models perform on separate tasks according to how they define their categories. This fine-grained breakdown allows the practitioner to select models that work well for their particular use case.
  \item Our benchmark creation methodology encourages more diversity and transparency to the practitioner compared to other alternatives. In comparison to other baselines like Alpaca-Eval and Arena-Hard v0.1, our benchmark has 84\% separability, 84\% agreement with confidence interval (95\%) with respect to Chatbot Arena rankings, 0.915 Spearman's correlation coefficient with respect to Chatbot Arena rankings and 0.04 Brier Loss Score.
  \item We also analyze the aforementioned metrics on our evaluation set with 4 LLM judges: GPT-4o\cite{openai2024gpt4technicalreport}, GPT-4o-mini \cite{OpenAI2024GPT4OMini}, Llama 3.1 405B Instruct and Llama 3.1 70B Instruct \cite{dubey2024llama3herdmodels}. Our overall findings suggest that while open-source models can be used to separate between model rankings, agreement with Chatbot Arena model rankings is roughly 10\% (405B) and 20\% (70B) than GPT-4o.
\end{itemize}
\section{Related Work}
At their core, benchmarks are tool to estimate LLM capabilities. There are many different flavors of benchmarks, spanning either across domains or various tasks. Some popular benchmarks include: Boolq \cite{clark2019boolqexploringsurprisingdifficulty}, MMLU \cite{hendrycks2021measuringmassivemultitasklanguage}, GSM8k \cite{cobbe2021trainingverifierssolvemath}, MATH \cite{hendrycks2021measuringmathematicalproblemsolving}, XSUM \cite{narayan2018dontdetailsjustsummary}, Hellaswag \cite{zellers2019hellaswagmachinereallyfinish}, and MGSM \cite{shi2022languagemodelsmultilingualchainofthought}. An expanded framework of static benchmark is AutoBencher which automatically creates new benchmarks which finds holes in knowledge of current SOTA LLMs \cite{li2024autobenchercreatingsalientnovel}.

These types of benchmarks have ground-truth references and compare how closely the LLM's completion aligns with those references. An inherent limitation with static benchmarks is that they are hosted on the internet and thus are susceptible test leakage contamination \cite{sainz-etal-2023-nlp, yang2023rethinkingbenchmarkcontaminationlanguage}. The other style of benchmarking relies on constructing a human evaluation trials on a set of evaluation prompts. Due to the expensive nature of human evaluation, a recent, cheaper alternative is to use SOTA LLMs to evaluate model completions either through single score or pairwise comparison with a reference answer, popularly referred to as LLM-as-a-Judge \cite{alpaca_eval, zheng2023judgingllmasajudgemtbenchchatbot, li2024crowdsourced, dubois2024alpacafarmsimulationframeworkmethods, verga2024replacingjudgesjuriesevaluating}.

This motivates the need for "live, refreshable" benchmarks so that the integrity of the benchmark can be maintained. LiveBench is a framework which sources data from arXiv papers, news articles, and datasets to periodically replace the stale prompts \cite{white2024livebenchchallengingcontaminationfreellm}. Chatbot arena is an open platform that allows online users to send prompts to two different models and compare/contrast the models' response \cite{chiang2024chatbotarenaopenplatform}. Users can then vote on which completion was superior. Other live benchmarks include DynaBench \cite{kiela2021dynabenchrethinkingbenchmarkingnlp}, LiveCodeBench \cite{jain2024livecodebenchholisticcontaminationfree}, and R2E \cite{jain2024r2e}. Our work lies in the intersection between LLM-as-a-Judge and live benchmarks as our data pipeline enables periodic refreshing of the evaluation set from existing clusters. Furthermore, our data pipeline is fairly general as it can consume a variety of diverse datasets (relative to Arena-Hard v0.1 and Alpaca-Eval), consists of using open-source models, and is flexible enough to work on the user's desired data.
\begin{figure*}[t]
    \centering
    \includegraphics[width=0.7\linewidth]{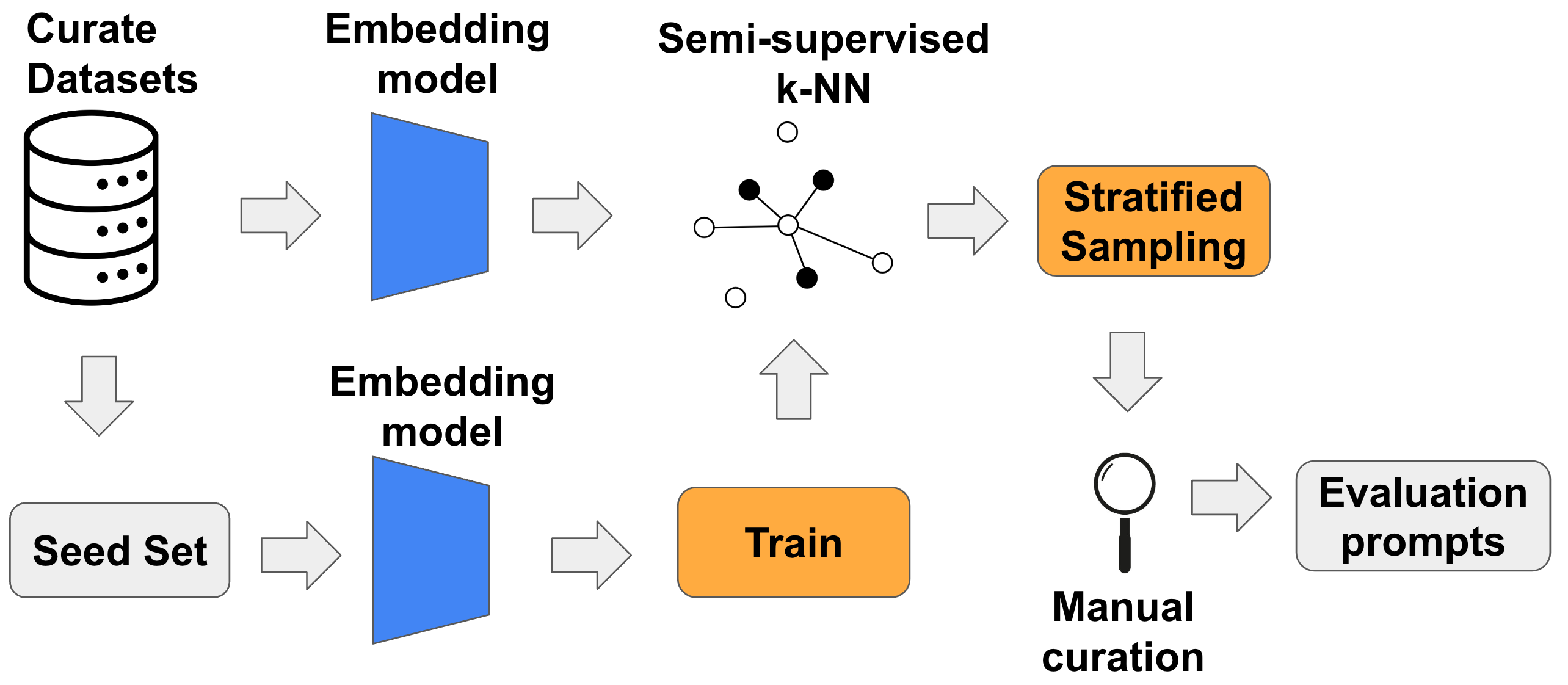}
    \caption{\textbf{Data pipeline}: After aggregating the prompts from datasets, we generate embeddings using a text embedding model. We set aside a set of prompts to use as a seed set for training the k-NN, label them into each category we care about, and generate their corresponding embeddings to train the k-NN with the embedding model. Subsequently, we classify the unlabeled data with our trained k-NN to create clusters of categories. We balance the clusters with stratified sampling and then manually curate the remaining prompts by removing overly long prompts (greater than 5000 words) and checking for low-quality content (nonsense prompts, NSFW etc.) to obtain the final evaluation set.}
    \label{fig:data_pipeline}
\end{figure*}

\section{Methodology}
In this section, we describe our approach to creating novel evaluation set using LLM-as-a judge. We enumerate the datasets that we source from to create our unlabeled corpus and subsequently describe our data pipeline for generating the evaluation set.

\subsection{Data Sources}
We use data sources from a variety of source to ensure we cover a variety of domains as well as languages. The domains we target can be broadly classified as the following: medical, law, finance, mathematics and coding. The languages we cover are standard but also more esoteric: Japanese (ja), Arabic (ar), Thai (th), Hungarian (hu), Russian (ru), Serbian (sr), Slovenian (sl), and Turkish (tr). Prompts that don't neatly fit into these groups fall into a catch-all general category. A complete list of all the data we use can be found in Table \ref{tab:Dataset Sources} in the Appendix.

\subsection{Data pipeline}
Our data pipeline can be divided into 3 distinct steps, as shown in Figure \ref{fig:data_pipeline}. We first take the data corpus and use an embedding model to generate their corresponding embedding. Each embedding encapsulates some level of semantic understanding of its associated prompt, and nearby embeddings typically encode similar semantic information. 

To generate the labels for the unlabeled data, we take inspiration from semi-supervised learning \cite{Hady2013}. We manually define a set of categories, curate a seed set of prompts which fall into those categories (assigning them distinct labels) and embed those prompts with the aforementioned embedding model. We train a $k$-NN model \cite{Mucherino2009} on top of those embeddings and use the $k$-NN to label the larger unlabeled corpus. 

The final step in our pipeline involves applying stratified sampling \cite{inbook} to each cluster. The reason for this last step is that we want our evaluation set to retain diversity of our larger data corpus rather than uniform random sampling. For each category, we sub-sample $100$ prompts from the aggregate clusters  and disregard clusters which have a lower count than the number of prompts we sampled. To obtain our final evaluation set, we manually curate the remaining prompts to ensure high quality, varied task capability and data diversity.

\section{Experimental Setup}
In this section, we discuss finer details about the data pipeline we mentioned in the prior section. experimental setup on a set of ten highly rated models\footnote{gpt-4o-2024-05-13, claude-3-5-sonnet-20240620, claude-3-opus-20240229, gemini-1.5-flash-latest, google/gemma-2-27b-it, Meta-Llama-3-70B-Instruct, claude-3-sonnet-20240229, Qwen/Qwen2-72B-Instruct, Meta-Llama-3-8B-Instruct, Mixtral-8x7B-Instruct-v0.1} as well as defining the metrics which determine the quality of the benchmark.
\subsection{Data pipeline details}

For the data pipeline, we use semi-supervised learning via a $k$-NN classifier.  We consider $13$ categories comprising of domains: finance, law, medical, maths, coding and languages: Arabic, Russian, Serbian, Hungarian, Japanese, Thai and Slovenian. We follow usual supervised training and via hyperparameter sweep over validation set yield $k=40$ as the best value of $k$.

To generate the embeddings of the unlabeled data collected, we use the e5-mistral-7b-instruct embedding model\cite{wang2024improvingtextembeddingslarge} for its strong performance on the Massive Text Embedding Benchmark (MTEB) Leaderboard \cite{muennighoff2022mteb} and multilingual capability.  If the $k$-NN encounters a sample which it is not familiar with or uncertain to label, we want those samples to be classified as general prompts. We use entropy of $k$-NN classifier probabilities of various categories for a given prompt as the measure of uncertainty. If entropy if too high entropy of the output of the classifier is too high, we bucket the sample into the default/general category \cite{settles2010survey}. We set the entropy threshold to be $1.5$ based on careful error analysis on the validation set. 
 
After labeling with $k$-NN, we conducted stratified sampling within each cluster, selecting $100$ samples for curation. We then filtered out excessively long prompts (longer than $5000$ words) that could overwhelm the judge's context window. Additionally, we reviewed the remaining prompts to eliminate those that were nonsensical or of low quality. During the evaluation, we observed that categories with a small number of examples had a significant impact on the category's win rate. The inherent variability of the LLM-as-a-Judge evaluation, even with a fixed random seed and temperature set to 0.0, made it challenging to discern which model performed better in those categories. To mitigate this uncertainty, we ensured that any category with fewer than 90-100 examples was supplemented with additional data, enabling us to obtain meaningful and interpretable results. Our final evaluation set comprises $1573$ examples.

\subsection{LLM-as-a-Judge Details}
We follow a similar scoring setup as Arena-Hard \cite{li2024crowdsourced} and Alpaca-Eval \cite{dubois2024lengthcontrolledalpacaevalsimpleway} where we use GPT-4o as a judge model and GPT-4o as a reference model as well. For each model we want to test, we obtain the completions and ask GPT-4o to record which model responses is better for the input prompt. In order to mitigate positional bias, we swap the completions between the model we are evaluating and the reference on a coin flip. 

For the judge prompt, we used the default prompt from the MT-Bench work with one notable change \cite{zheng2023judgingllmasajudgemtbenchchatbot}. When we evaluated multilingual prompts with LLM-as-a-judge, the judge at times incorrectly awards wins to models which don't necessarily follow instructions. Given the sentence "Please respond 'How does the economy work?' in Hungarian," two models might respond differently: 1) one provides a detailed English response with bulleted lists, while 2) the other responds concisely in Hungarian. The judge model will rate the model answering in the incorrect language higher, which is clearly not a measure of the model's multilingual capability \cite{marchisio2024understandingmitigatinglanguageconfusion}. In order to reduce these incorrect decisions, we modified the judge prompt to specifically penalize responses that respond to the prompt in the incorrect language. 

In addition to issues with multilingual queries, we also note specifically for coding that GPT-4o seems to prefer models which provide detailed explanations to the code even if the code provided is of lower quality compared to a model which has better code quality but is not as verbose. This leads to scenarios where models that have  chat  but lower benchmark performance (e.g. HumanEval \cite{chen2021evaluating}) obtain higher winrate than models which are objectively better on coding prompts. To circumvent this issue, we explicitly prompt GPT-4o that it should focus on the correctness of the response as opposed to the style of the response. Our judge template can be found in the Appendix.

\subsection{Obtaining Confidence Intervals}
We follow the setup outlined in \textit{Li et. al} \cite{li2024crowdsourced, chiang2024chatbotarenaopenplatform}. We use the Bradley-Terry model in order to model the preference distribution between models on the leaderboard and the reference model (GPT-4o in our case). We aggregate preference pairs between models and perform 100 rounds of bootstrapping to obtain 95\% confidence intervals for each model ranking. 

We conduct the same analysis with annotations, denoting for each prompt which model response was preferred, from the Alpaca-Eval repo to obtain mean ELO rankings and 95\% confidence intervals according to their leaderboard. Since similar artifacts (model preference comparisons) are not updated on Arena-Hard v0.1, we take the model winrates (ELO scores not listed) and 95\% confidence intervals from their repo\footnote{7/26/2024}. For Chatbot Arena, we do the same thing and took model winrates/ELO scores as well as the confidence intervals from the website\footnote{7/25/2024} as a source of ground truth.

\begin{table*}[!htb]
\centering
\begin{tabular}{ccccc}
\hline
 & Chatbot Arena & Arena Hard v0.1 & Alpaca-Eval 2.0 LC & Ours \\
\hline
Separability & \textbf{100\%} & 80\% & 73.33\% & 84.44\% \\
Agreement with CI (95\%) & N/A & 75.50\% & 64.44\% & \textbf{84.44\%} \\
Spearman's Correlation & N/A & 0.187 & 0.2969 & \textbf{0.915} \\
Brier Score & N/A & N/A & 0.0937 & \textbf{0.0417} \\
\hline
\end{tabular}
\caption{Main results comparing the various benchmarks.}
\label{table:Main results}
\end{table*}

\subsection{Metrics}
There are four different metrics we use to judge the efficacy of a benchmark. The first of these is Spearman's correlation coefficient, which measures the rankings order between the two benchmarks. The other metrics are: separability, agreement with Confidence Interval (CI), and Brier Score. Separability refers to how well the benchmark can separate various models with high confidence. In particular, if  on benchmark A model M1 has a higher ELO/winrate than model M2 and $C_{M}$ refers to the confidence intervals of model M, $S$ is a binary variable indicating if benchmark A is able to separate between model M1 and M2, $S = \mathbf{1}_{C_{M1} \cap C_{M_2} = \emptyset}$. The separability is then calculated as a ratio over all possible model pairs. Agreement with CI measures how well benchmarks A and B confidently distinguish between two models with the same ordering. The Brier Score evaluates an LLM benchmark's ability to predict the ranking of a pair of competing models, rewarding confidence in accurate predictions and penalizing confidence in incorrect ones. More details behind these metrics can be found in \cite{li2024crowdsourced}. Ultimately, we want our benchmark to align with Chatbot Arena as that is seen as an oracle for modeling human preferences.

\section{Results}
\subsection{Separability, Agreement with CI (95\%), Pair Brier Score}
Our main results can be found in Table \ref{table:Main results}. With the exception of Chatbot Arena, our benchmark's separability is 84.4\% compared to other baselines like Arena-Hard v0.1 (80\%) and Alpaca-Eval 2.0 LC (73.33\%), which shows that our benchmark can better differentiate amongst different models.

One interesting datapoint regarding separability is Chatbot Arena's score of 100\% which may be attributed to a combination of two factors: 1) Chatbot Arena has more battles than any of the benchmarks listed in Table \ref{table:Main results} and 2) Chatbot Arena includes battles between many different models rather than fixing a reference model like the other benchmarks. By providing the Bradley-Terry model bootstrapping process with more varied battles, Chatbot Arena is able to produce tighter confidence intervals, suggesting a future avenue for investigation is whether confidence estimation should include multiple reference answers during judging to more closely simulate Chatbot Arena.

Our benchmark showed an 84.44\% agreement with CI with respect to Chatbot Arena, which is higher than Arena-Hard v0.1's 75.50\% and Alpaca-Eval 2.0 LC's 64.44\%. This demonstrates that our benchmark has higher alignment with respect to Chatbot Arena which is supposed to be approximation of human preferences. In addition, our benchmark has a Spearman’s correlation coefficient of 0.915, indicating a strong correlation in rankings order compared to Alpaca-Eval 2.0 LC's 0.2969. While our leaderboard ranking consists of 10 models, the pool of models we have included are the latest SOTA models that have been released so as to have the maximum amount of overlap possible. Finally, our benchmark scored a Brier score of 0.0417, which is lower than Alpaca-Eval 2.0 LC's 0.0937, demonstrating better confidence in accurate predictions.

\subsection{Diversity}
Due to our data sources being quite diverse rather than simply just ChatBot Arena \cite{chiang2024chatbotarenaopenplatform}, we are able to have more diversity in our evaluation set. To demonstrate this, we label Arena-Hard v0.1 with our kNN model using the entropy threshold to get a distribution of categories in that evaluation set. As shown in Figure \ref{fig:arena-hard_v0.1_pie_chart}, there is an over-representation of coding prompts, which comes from a byproduct of their data pipeline filtering for the hardest, highest quality which skews towards coding. Similarly, Alpaca-Eval's prompt distribution shown in  Figure \ref{fig:alpaca_eval_pie_chart} demonstrates that there is a large emphasis on general chat queries, along with some coding and math prompts while medical and law prompts are relatively underrepresented.

\begin{figure*}
    \centering
    \subfigure[Our evaluation set category breakdown]{
        \includegraphics[width=0.46\textwidth]{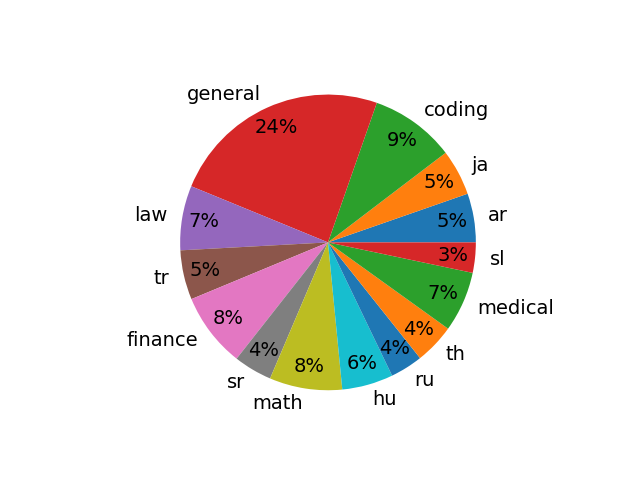}
        \label{fig:private_eval_breakdown}
    }
    \hspace{0.1cm}
    \subfigure[LMSys Conversations category breakdown]{
        \includegraphics[width=0.46\textwidth]{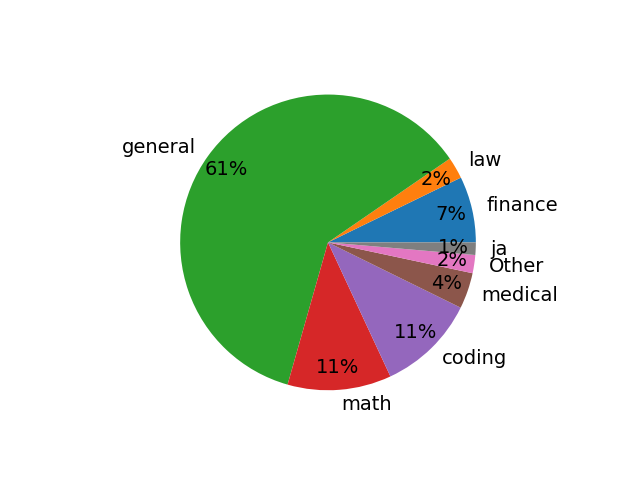}
        \label{fig:lm_sys_category_breakdown}
    }
    \caption{We look at the category breakdown on our evaluation set (Figure \ref{fig:private_eval_breakdown}) compared to LM-SYS Conversations (Figure \ref{fig:lm_sys_category_breakdown}). We can clearly see that our evaluation set covers more languages and niche domains such as law and medical categories are a higher percentage of our evaluation set.}
    \label{fig:pie_chart}
\end{figure*}

Our evaluation set breakdown in Figure \ref{fig:private_eval_breakdown} which covers more domains than the baseline, such languages like Arabic, Japanese, Hungarian and more. The close to equal distribution amongst the categories is likely due to the effect stratified sampling. We compare how our evaluation set category breakdown compared with LM-SYS Conversations (using our k-NN labeling approach) \cite{zheng2023judging} in Figure \ref{fig:lm_sys_category_breakdown}, which is a snapshot of cleaned Chatbot Arena conversations from April to June 2023. In Figure \ref{fig:lm_sys_category_breakdown}, "Other" refers to the languages our k-NN classifier recognizes but groups them together collectively. We note that this distribution looks similar to Alpaca-Eval and the general category may contain additional languages not recognized by the classifier so it may have exceeded the entropy threshold.

\subsection{Category Separability}
Due to our unique ability to categorize the prompts, we can compute category separability for all the various categories in our evaluation set. Across 14 different categories, we do the same bootstrapping procedure on the category data to obtain the mean winrate/ELO and 95\% CI, shown in Table \ref{table:winrate_elo}. In general, there is a drop in separability when we look both at ELO ratings and winrate due to each category having a lower number of samples and thus larger CIs as a result.

The category-wise separability can act as an indicator which categories are superior at testing out the performance of models. Interestingly, across ELO and winrate rankings, Hungarian has the best separability of all categories, achieving 66.67\% and 75.56\% respectively. The medical category seems to be lowest separability around 55.56\% and 68.89\% respectively. The separability also indicates to use which categories we may need to add more samples to improve the confidence intervals.

\begin{table}
\centering
\begin{tabular}{l|c|c}
Category & Ranking winrate & Ranking ELO \\
\hline
ar       & 73.33\% & 57.78\% \\
ru       & 71.11\% & 55.56\% \\
finance  & 75.56\% & 57.78\% \\
sr       & 71.11\% & 53.33\% \\
tr       & 73.33\% & 55.56\% \\
general  & 77.78\% & 55.78\% \\
hu       & 75.56\% & 66.67\% \\
ja       & 71.11\% & 57.78\% \\
medical  & 68.89\% & 55.56\% \\
law      & 73.33\% & 51.11\% \\
th       & 71.11\% & 57.78\% \\
coding   & 73.33\% & 55.56\% \\
sl       & 77.78\% & 53.33\% \\
math     & 73.33\% & 55.56\% \\
\end{tabular}
\caption{Winrate and ELO separability for different categories}
\label{table:winrate_elo}
\end{table}

\subsection{Using different judges}

\begin{table*}
\centering
\begin{tabular}{ccccc}
\hline
 & GPT-4o & GPT-4o-mini & Llama 3.1 405B & Llama 3.1 70B \\
\hline
Separability & 84.44\% & 82.22\% & 82.22\% & 84.44\% \\
Agreement with CI (95\%) & 84.44\% & 76.77\% & 75.55\% & 66.66\% \\
Spearman's Correlation & 0.915 & 0.0787 & 0.0787 & 0.0787 \\
Brier Score & 0.0417 & 0.062 & 0.0603 & 0.0955 \\
\hline
\end{tabular}
\caption{Comparing various judges on our evaluation set.}
\label{table:Judge_table}
\end{table*}

We conduct an ablation of judge models on our evaluation, as we want to understand the effect of judge models on separability, Agreement with CI (95\%) and Brier Score. We consider GPT-4o mini as one of the judges to be a small-closed source foil to GPT-4o. The other judges that we consider are open source models such as: Llama 3.1 405B instruct (using SambaNova's developer API)\footnote{https://sambanova.ai/fast-api} and Llama 3.1 70B Instruct-Turbo\footnote{https://api.together.ai/models/meta-llama/Meta-Llama-3.1-70B-Instruct-Turbo}. We follow the same setup as gpt-4o with these other judge models. 

Our results are shown in Table \ref{table:Judge_table}. In terms of separability, GPT-4o-mini and 405B get 82.2\% and 70B get 84\% separability, comparable to GPT-4o's separability. 405B and GPT-4o-mini attain similar Agreement with CI (95\%) close to 76\% while 70B is almost 10 points lower; GPT-4o is the clear winner having the highest agreement with CI (95\%). With the exception of 70B, all models get similar Brier Scores indicating that the Bradley-Terry models used to generate the rankings on confidence intervals for each judge are similarly confident. 70B's high Brier score (relative to other judges), in addition to Agreement with CI, indicates that it poor judge than the other listed in Table \ref{table:Judge_table}. 

\begin{figure}[h]
    \centering
    \includegraphics[width=1.0\linewidth]{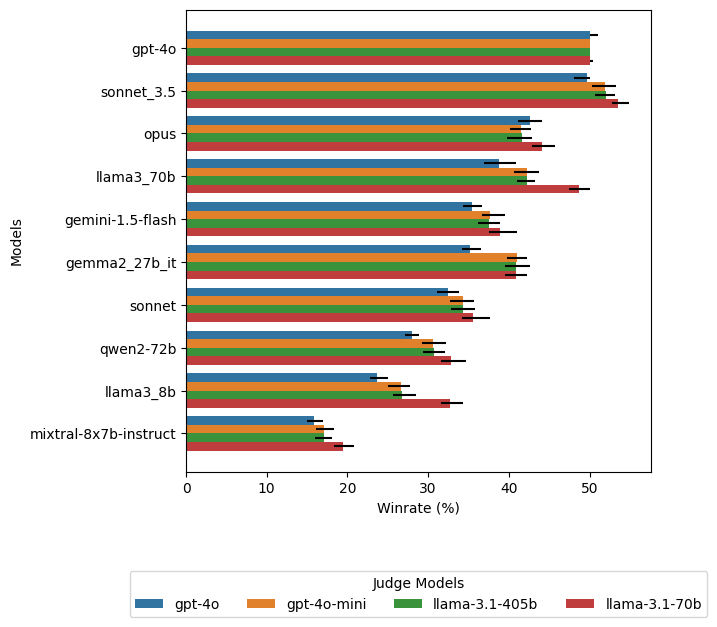}
    \caption{Visual comparison of different judge's separability on our benchmark.}
    \label{fig:judge_separability_barplot}
\end{figure}

The Spearman's correlation coefficient (with respect to ChatBot Arena rankings) seems to indicate that GPT-4o-mini, Llama 3.1 405B, and 70B are poor judges getting a correlation of only 0.0787 vs. GPT-4o's 0.915. Looking at Figure \ref{fig:judge_separability_barplot}, it seems this aberration comes from both judges rating Claude Sonnet 3.5 over GPT-4o, Llama 3 70b over Claude Opus and Gemma2 27B over Gemini 1.5 Flash. Of course, Spearman's correlation only measures correlation the final rank order of models with respect to ChatBot Arena and is a strictly weaker metric than Agreement with CI (95\%). This finding seems to suggest while weaker closed-source models (like GPT-4o-mini) and open-source judge models seem to be able to separate other models based on capability, they still lack the preciseness that GPT-4o offers to align with rankings from Chatbot Arena.

\section{Limitations/Future Work}
There are certain limitations to our work. Currently, the categories we enumerate in our data pipeline is manually specified by humans and significant curation is done to ensure high quality prompts; for future work, we want to expand to using LLMs as category generators as well as quality checkers to automate the human effort out of this pipeline. For improving our leaderboard, we wish to add more models to be more representative of the entire spectrum of other leaderboards and futhur increasing the quality of the Bradley-Terry models we use to obtain the model's confidence intervals. In order to improve category separability, we look to creating a methodology on figuring out the minimum number of samples required to improve separability.

The other aspect of future work relies to details regarding LLM-as-a-judge evaluation. Typically, the judge models are ablated but less explored is the quality of the reference answer and whether one can use a weaker model instead of a stronger one to see if metrics are maintained. Current metrics define how separable a benchmark is and how much it aligns with human preferences but fails to account for the composition and diversity of the underlying data. For future work, we seek to quantify the diversity of each benchmark to understand how many capabilities/domains it spans.
\section{Conclusion}
We introduce a data pipeline that leverages via semi-supervised learning with a k-NN to enable practitioners to create benchmarks on their own data for targeted domains. Through evaluations of ten various closed and open-sourced models, we demonstrated that our benchmark achieves higher separability and agreement with CI with respect to Chatbot Arena, nearly 5 and 10 percentage points higher than the next best baseline, respectively. Our benchmark covers a wide variety of topics such as finance, medicine, legal and different languages absent in other LLM as a judge benchmarks. We hope that LLM developers can use our data pipeline to create their own benchmarks to evaluate their models for their particular use-case. 

\bibliography{references}
\bibliographystyle{icml2024}
% \newpage
\onecolumn
\section{Appendix}
\label{appendix}
\subsection{Data Sources}
\begin{table}[htb]
    \centering
    \begin{tabular}{c}
         Datasets \\
         \hline
         LMSys Chatbot Arena \cite{chiang2024chatbotarenaopenplatform} \\
         PubMedQA \cite{jin2019pubmedqa} \\
         MathQA \cite{amini-etal-2019-mathqa} \\
         No Robots \cite{no_robots} \\
         Aya \cite{singh2024ayadatasetopenaccesscollection} \\
         Legal reddit \cite{li-etal-2022-parameter} \\
         Legal Summ. BillSum \cite{kornilova-eidelman-2019-billsum} \\
         Airoboros-gpt4 \cite{jondurbin_2024} \\
         Finance Advisor \cite{gaurang_bharti_2024} \\
         Finance Bier QA \cite{thakur2021beir} \\
         MMLU \cite{hendrycks2021measuringmassivemultitasklanguage} \\
         TruthfulQA \cite{lin2022truthfulqameasuringmodelsmimic} \\
         GSM8K \cite{cobbe2021trainingverifierssolvemath} \\
         \hline
    \end{tabular}
    \caption{Dataset Sources used as input to the data pipeline in Figure \ref{fig:data_pipeline}.}
    \label{tab:Dataset Sources}
\end{table}

Table \ref{tab:Dataset Sources} includes various datasets across multiple domains such as medical, legal, financial, and multilingual categories. These sources were selected to ensure a wide range of coverage, contributing to the diversity of the evaluation set. The datasets listed here were crucial for constructing the domain-specific evaluation sets, allowing for the thorough testing of models across different contexts and languages.

\subsection{Judge Template}
Below is our judge template that we used for our LLM-as-a-judge evaluation:

\textit{Please act as an impartial judge and evaluate the quality of the responses provided by two AI assistants to the user question displayed below. You should choose the assistant that follows the user's instructions and answers the user's question better, as well as answering in the desired language of the user. Your evaluation should consider factors such as the helpfulness, relevance, accuracy, depth, creativity, and level of detail of their responses. Begin your evaluation by comparing the two responses and provide a short explanation. Avoid any position biases and ensure that the order in which the responses were presented does not influence your decision. Do not allow the length of the responses to influence your evaluation. Do not favor certain names of the assistants. Be as objective as possible. Your evaluation should only focus on the correctness of the response. After providing your explanation, output your final verdict by strictly following this format: [[A]] if assistant A is better, [[B]] if assistant B is better, and [[C]] for a tie.}

\subsection{Evaluation Tool}
\begin{figure}[htb]
    \centering
    \includegraphics[width=1.0\linewidth]{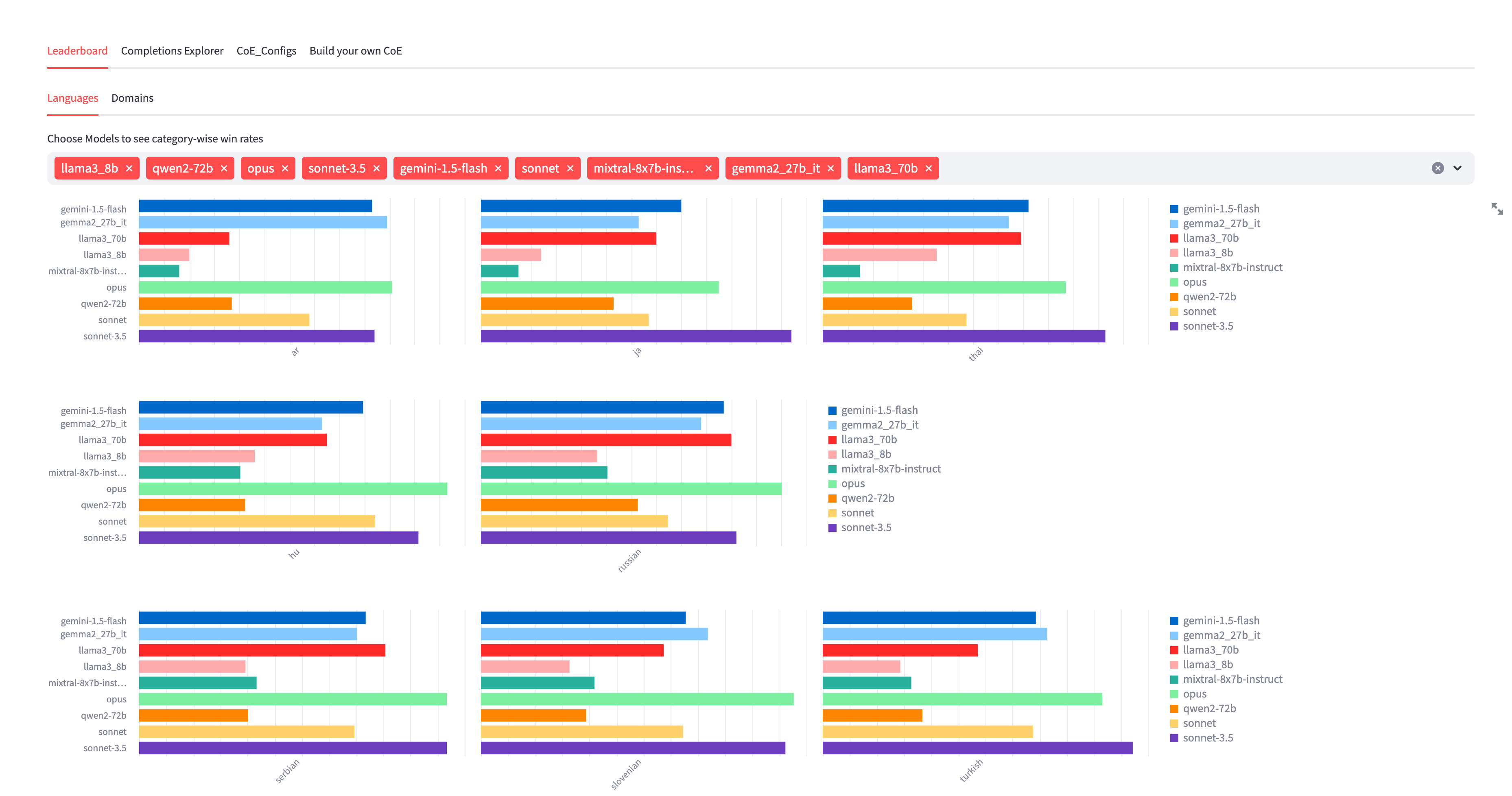}
    \caption{A screenshot of our evaluation tool.}
    \label{fig:evaluation tool}
\end{figure}

With the notion of self-defined categories and using the LLM-as-a-judge framework, we create an evaluation tool which loads an internal leaderboard from a csv file and breaks down the winrate into several categories the user defined. The UI shows the leaderboard in a dataframe and shows the winrates in set of bar plots across different categories. A screenshot of the tool can be seen in Figure \ref{fig:evaluation tool}.

There is also a feature which enables the user to view completions on the evaluation from both the model the user is interested in, the reference model, and the judge model to examine its reasoning. This tool enables the user to examine where the model they are developing is performing better than other competitors and areas where improvement is required.

% \newpage
% \appendix
% \onecolumn
% \section{You \emph{can} have an appendix here.}

\end{document}